\documentclass[logo, copyright, nonumbering]{style}

\usepackage[utf8]{inputenc} %
\usepackage[T1]{fontenc}    %
\usepackage{hyperref}       %
\usepackage{url}            %
\usepackage{booktabs}       %
\usepackage{amsfonts}       %
\usepackage{nicefrac}       %
\usepackage{microtype}      %
\usepackage{xcolor}         %
\usepackage{xspace}

\usepackage{graphicx}
\usepackage{bm}
\usepackage{amsmath}
\usepackage{amssymb}
\usepackage{todonotes}
\usepackage{pifont}
\usepackage{multirow}
\usepackage{multicol}
\usepackage{wrapfig}
\usepackage{colortbl}
\usepackage{caption}
\usepackage{bbm}
\usepackage{tabularx}
\usepackage{etoolbox}
\usepackage[most, listings]{tcolorbox}
\usepackage{fancyvrb}

\newcommand{\algname}{KT$^2$\xspace}

\DeclareMathOperator*{\argmax}{arg\,max}

\makeatletter
\def\thanks#1{%
    \footnotemark%
    \protected@xdef\@thanks{\@thanks
        \protect\footnotetext[\the\c@footnote]{#1}}%
}
\def\@fnsymbol#1{\ensuremath{\ifcase#1\or \dagger\or *\or \ddagger\or 
   \mathsection\or \mathparagraph\or \|\or **\or \dagger\dagger 
   \or \ddagger\ddagger \else\@ctrerr\fi}}

\renewcommand\AB@authnote[1]{}  %

\newcolumntype{Y}{>{\raggedright\arraybackslash}X}
\title{A Hierarchical Probabilistic Framework for Incremental Knowledge Tracing in Classroom Settings}
\runningtitle{Knowledge-Tree-based Knowledge Tracing}

\newcommand{\aspace}{\hspace{1em}}
\newcommand{\ucsb}{$^{1}$}
\newcommand{\ibm}{$^{2}$}
\newcommand{\umich}{$^{3}$}
\newcommand{\harvard}{$^{4}$}
\newcommand{\equal}{$^{\dagger}$}

\author{
\textbf{Xinyi Gao}\ucsb \thanks{Equal contribution.} \aspace 
\textbf{Qiucheng Wu}\ucsb\equal \aspace 
\textbf{Yang Zhang}\ibm\equal \\
\textbf{Xuechen Liu}\umich \aspace 
\textbf{Kaizhi Qian}\ibm \aspace 
\textbf{Ying Xu}\harvard \aspace 
\textbf{Shiyu Chang}\ucsb
}

\affil[1]{UC Santa Barbara}
\affil[2]{MIT-IBM Watson AI Lab}
\affil[3]{University of Michigan}
\affil[4]{Harvard University}

\renewcommand{\emailsdata}{
  \texttt{xinyigao@ucsb.edu, qiucheng@ucsb.edu}
}

\begin{document}

\newcommand{\fullplan}{\textsc{FullPlan}}
\newcommand{\fullplansp}{\textsc{FullPlan}~}
\newcommand{\replan}{\textsc{Replan}}
\newcommand{\replansp}{\textsc{Replan}~}
\newcommand{\boxnet}{\texttt{BoxNet}}
\newcommand{\boxnetsp}{\texttt{BoxNet}~}
\newcommand{\boxnettwod}{\texttt{BoxNet2D}}
\newcommand{\boxnettwodsp}{\texttt{BoxNet2D}~}
\newcommand{\boxnetthreed}{\texttt{BoxNet3D}}
\newcommand{\boxnetthreedsp}{\texttt{BoxNet3D}~}

\newcommand{\hlg}[2]{\setlength{\fboxsep}{0.3pt}\colorbox{green!#2}{\rule[-.05\baselineskip]{0pt}{.7\baselineskip}{#1}}}
\newcommand{\hlr}[2]{\setlength{\fboxsep}{0.3pt}\colorbox{red!#2}{\rule[-.05\baselineskip]{0pt}{.7\baselineskip}{#1}}}
\newcommand{\hlb}[2]{\setlength{\fboxsep}{0.3pt}\colorbox{aqua!#2}{\rule[-.05\baselineskip]{0pt}{.7\baselineskip}{#1}}}
\definecolor{grey}{rgb}{0.8,0.8,0.8}
\definecolor{aqua}{rgb}{0, 1, 1}
\definecolor{steel}{rgb}{0.2734, 0.5078, 0.7031}
\definecolor{slate}{rgb}{0.1836, 0.3086, 0.3086}
\definecolor{tableheadcolor}{RGB}{200,200,200}
\definecolor{transgray}{gray}{0.9}
\definecolor{lightblue}{rgb}{0.18,0.39,0.62}
\definecolor{blue2}{rgb}{0.1,0.5,0.65}
\definecolor{green2}{RGB}{0, 100, 0}
\definecolor{pink}{rgb}{0.8,0.4,0.4}
\definecolor{darkred}{RGB}{165, 42, 42}
\definecolor{lightbluebg}{rgb}{0.96,0.98,1.0}
\definecolor{rulegray}{rgb}{0.7,0.7,0.8}

\definecolor{orange2}{RGB}{221, 132, 82}
\definecolor{red2}{RGB}{196, 78, 82}
\definecolor{purple2}{RGB}{149, 108, 180}
\definecolor{darkblue2}{RGB}{76, 114, 176}
\definecolor{lightblue2}{RGB}{100, 181, 205}
\newcommand{\myrowcolour}{\rowcolor{tableheadcolor}}

\newcommand{\website}{\url{https://github.com/UCSB-NLP-Chang/KT2}\xspace}
\maketitle

\begin{abstract}

Knowledge tracing (KT) aims to estimate a student's evolving knowledge state and predict their performance on new exercises based on performance history. Many realistic classroom settings for KT are typically low-resource in data and require online updates as students' exercise history grows, which creates significant challenges for existing KT approaches. To restore strong performance under low-resource conditions, we revisit the hierarchical knowledge concept (KC) information, which is typically available in many classroom settings and can provide strong prior when data are sparse. We therefore propose Knowledge-Tree-based Knowledge Tracing (\algname), a probabilistic KT framework that models student understanding over a tree-structured hierarchy of knowledge concepts using a Hidden Markov Tree Model. \algname estimates student mastery via an EM algorithm and supports personalized prediction through an incremental update mechanism as new responses arrive. Our experiments show that \algname consistently outperforms strong baselines in realistic online, low-resource settings.
\iconbar{
    \href{https://github.com/UCSB-NLP-Chang/KT2}{\faGithub~GitHub}\quad
}

\end{abstract}

\section{Introduction}
\label{sec:intro}

Knowledge tracing (KT) refers to an educational task that, given historical exercises performance of a set of students, aims to predict whether they can solve a new exercise correctly. It can be regarded as modeling the evolution of a student's knowledge during learning, which is essential in personalized learning systems, enabling dynamic adaptation to students' needs. While prior works on knowledge tracing impose very loose constraints about the availability of data \cite{DKT, AKT}, in many practical real-world classroom scenarios, these constraints tend to be much more stringent, due to various privacy and usability considerations. In particular, three practical constraints have usually been understudied.

\noindent $\bullet$ \textbf{Cold Start.} For each target student, existing works usually assume that abundant historical data from that student is available before the prediction starts. This implies that the prediction would need to happen at a very late stage of the student's learning -- only after many exercises have been done. To make KT meaningful, the prediction should start shortly after the target student starts learning, with only a few exercises completed. This creates a cold start scenario for KT.

\noindent $\bullet$ \textbf{Online Update.} Conventional evaluation of KT methods usually involves only one-time prediction. However, in practice, new student exercises constantly arrive in a streaming fashion, requiring KT methods to be efficiently updated to capture the evolution of students' understanding.

\noindent $\bullet$ \textbf{Limited Peers.} Many KT methods require a large number of peers to form a sufficiently large historical dataset as the training set. In practice, the number of peers may be limited due to privacy protection.

The aforementioned constraints create significant challenges for existing KT algorithms. Specifically, existing KT approaches can be broadly categorized into two types. The first type, \textit{the deep-learning-based methods} \cite{DKVMN, AKT, SAKT, simpleKT}, trains neural architectures to model complex student performance patterns. Although these methods have shown strong performance under benchmark settings, they require large amounts of training data to achieve good performance, which is unavailable under the practical constraints. 

On the other hand, the second type of methods, the \textit{large language model (LLM)-based methods} \cite{FSP,FT-LLM}, leverages LLMs to perform the prediction tasks, either fine-tuning the LLM or selecting a subset of the target student's historical data as in-context examples. While these methods have improved data efficiency and online flexibility, they often struggle to capture fine-grained patterns in historical data and may make predictions based on naive decision rules.

In summary, there appears to be a fundamental trade-off between data availability and a model’s capacity to capture complex dependencies in student responses. Yet, one crucial source of information—long underused in existing KT methods—could help overcome this bottleneck: \textbf{knowledge concepts}.

Knowledge concepts (KCs) refer to the labeled topics, knowledge areas, or skills associated with each exercise. These concepts often follow a hierarchical structure, where broader concepts encompass finer-grained sub-concepts. KCs are readily available in many classroom settings, and can be surprisingly effective for modeling student learning. Intuitively, if a student has not mastered a particular concept, they are likely to answer related questions incorrectly. Moreover, their performance on conceptually adjacent or dependent concepts may also be affected. In low-resource classroom settings, these knowledge labels, along with their hierarchical organization, can provide strong structural priors that enhance prediction performance. \textbf{Can we leverage hierarchical knowledge concepts to enable data-efficient, flexible, and online knowledge tracing for realistic classroom settings?}

\begin{figure*}[t]
  \centering
  \includegraphics[width=\textwidth]{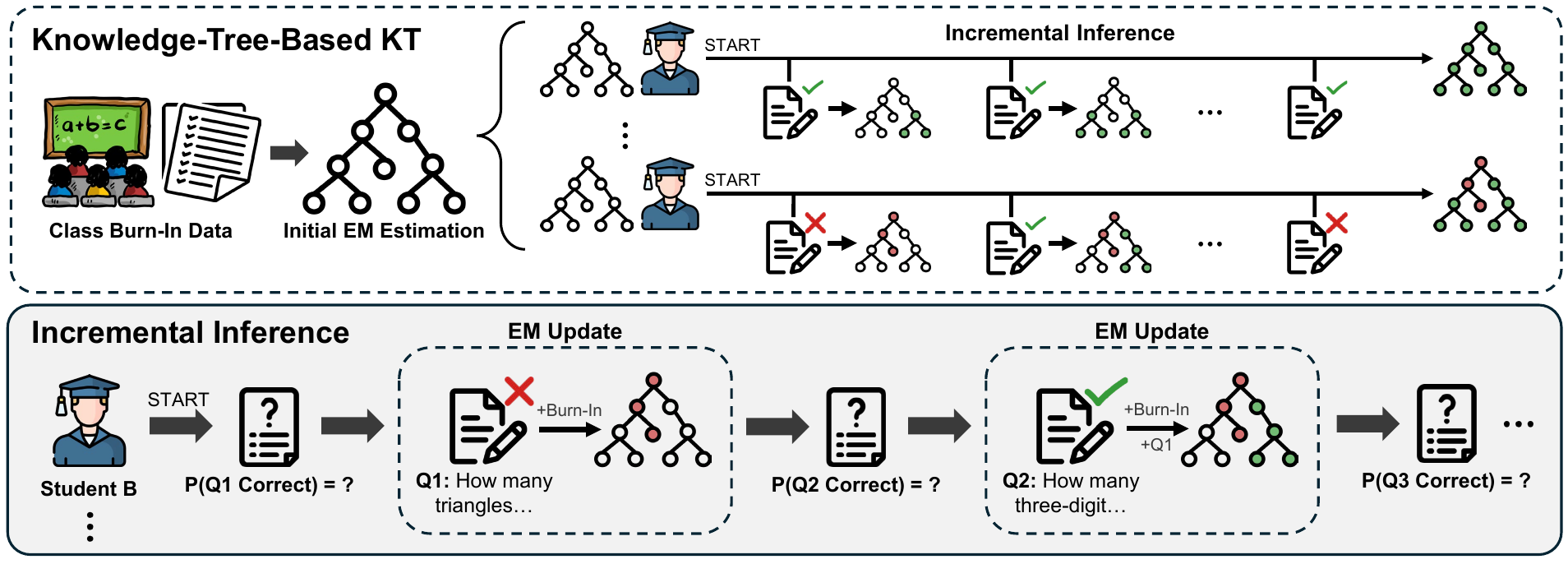}
  \vspace{-6mm}
  \caption{\algname framework. The model first performs a class-level EM estimation using all students' burn-in responses to initialize a global knowledge model. Each student is then assigned a personalized knowledge concept tree, which is incrementally updated with each new interaction via one-step EM updates during inference.}
  \label{fig:framework}
  \vspace{-2mm}
\end{figure*}

Motivated by this observation, we propose Knowledge-Tree-based Knowledge Tracing (\algname), a probabilistic framework for low-resource, online knowledge tracing. \algname builds a Hidden Markov Tree Model~\cite{HMT} based on the hierarchical tree structure of KCs, where the hidden variables represent the student’s mastery of each KC, and the observed variables correspond to their correctness on individual exercise questions. As shown in Fig.~\ref{fig:framework}, the model parameters are first estimated using a small amount of initial (``burn-in'') data via the standard Expectation-Maximization (EM) algorithm~\cite{EM}, and are then incrementally updated as new exercise responses are observed. \algname enables principled prediction of future performance of target students by computing the distribution of correctness on upcoming exercises variables, conditioned on the students' historical exercise performance.

Our experiments demonstrate that \algname effectively addresses the challenges of low-resource knowledge tracing. For instance, with only 50 target students and as few as five exercises per student, \algname consistently outperforms the strongest baselines. 
In the online setting, \algname requires only a single EM step to incorporate new data, enabling efficient and consistent performance updates.

\section{Method}

\subsection{Problem Formulation}

The problem of KT can be formulated as follows. Denote $\mathcal{I}$ as a set of student IDs. For each $i \in \mathcal{I}$, denote $\mathcal{N}_i$ as a set of exercise IDs for which student $i$ has solved. For each $i \in \mathcal{I}$, and $n \in \mathcal{N}_i$, define $Q_{ni}$ as a random binary variable:
\begin{equation}
    Q_{ni} = 
    \begin{cases}
        1 &  \mbox{ if student $i$ answers question $n$ correctly,}\\
        0 & \mbox{ otherwise.}
    \end{cases}
    .
\end{equation}

Finally, define 
\begin{equation}
    \mathcal{Q}_i= \{Q_{ni}: n \in \mathcal{N}_i\}
    \label{eq:observed_set}
\end{equation}
as a set of historical exercise variables of student $i$. Given a new question $n^*$, our goal is to predict:
\begin{equation}
    p_{\bm \theta}(Q_{n^*i} = 1 | \mathcal{Q}_i).
    \label{eq:pred}
\end{equation}

Here, $\bm \theta$ represents a set of parameters that defines the distribution, which will be estimated based on all the observed historical data, $\cup_i \mathcal{Q}_i$.

We here restate the real-world constraints mentioned in Sec.~\ref{sec:intro} formally:

$\bullet$ \textbf{Cold Start.} At the onset, each $\mathcal{Q}_i$ is a very small set, making the conditional probability in Eq.~\eqref{eq:pred} uninformative and the estimation of $\bm \theta$ challenging.

$\bullet$ \textbf{Online Update.} Each $\mathcal{Q}_i$ is expanding, requiring Eq.~\eqref{eq:pred} and $\bm \theta$ to be constantly updated.

$\bullet$ \textbf{Limited Peers.} $\mathcal{I}$ is a small set, adding to data scarcity when estimating $\bm \theta$.

In the following, we will discuss how the hierarchical KCs can alleviate the challenges.

\subsection{Knowledge Concept Tree}

Many hierarchical KC structures can be organized into trees. Fig.~\ref{fig:assumption}(left) shows a portion of an example knowledge concept tree, where parent nodes represent broader KCs and child nodes finer ones. Each edge represents an entailment relationship. We make two assumptions on our classroom setting: \ding{182} We have access to a knowledge concept tree that contains all the KCs covered in the exercises; and \ding{183} each exercise question is labeled with \textit{one KC at the leaf node.}

\subsection{The \algname Model}

\algname incorporates the knowledge concept tree information into the probabilistic modeling of the $Q_{ni}$ variables. Specifically, denote $\mathcal{C}$ as the set of all KCs. For each $i \in \mathcal{I}$, and $c \in \mathcal{C}$, define $K_{ci}$ as the following random binary variable:
\begin{equation}
    K_{ci} = 
    \begin{cases}
        1 & \mbox{if student $i$ masters KC $c$,} \\
        0 & \mbox{otherwise.}
    \end{cases}
    .
\end{equation}

\algname introduces a Hidden Markov Tree Model to model the joint probability distribution of $\{Q_{ni}\}$ (as observed variables) and $\{K_{ci}\}$ (as hidden variables) via the following assumptions.

\textbf{First}, all the variables are independent across different students, \emph{i.e.},
\begin{equation}
    p\big(\cup_{n,i} \{Q_{ni}\}, \cup_{c,i} \{K_{ci}\}\big) = \prod_i p\big(\cup_{n} \{Q_{ni}\}, \cup_{c} \{K_{ci}\}\big).
    \label{eq:indep}
\end{equation}

\textbf{Second}, for each student $i$, the corresponding probability distribution can be decomposed as
\begin{equation}
    \begin{aligned}
    & p\big(\cup_{n} \{Q_{ni}\}, \cup_{c} \{K_{ci}\}\big) \\
    = & \underbrace{p\big(\cup_{c} \{K_{ci}\}\big)}_\text{transition} \cdot \underbrace{p\big(\cup_{n} \{Q_{ni}\} | \cup_{c} \{K_{ci}\}\big)}_\text{emission},
    \end{aligned}
\end{equation}
where the first term, representing the \textit{transition probabilities}, models interdependencies among the mastery of different KCs; the second term, representing the \textit{emission probabilities}, models how the mastery of KCs indicates the exercise correctness.

\begin{figure}[t]
  \centering
  \includegraphics[width=0.48\textwidth]{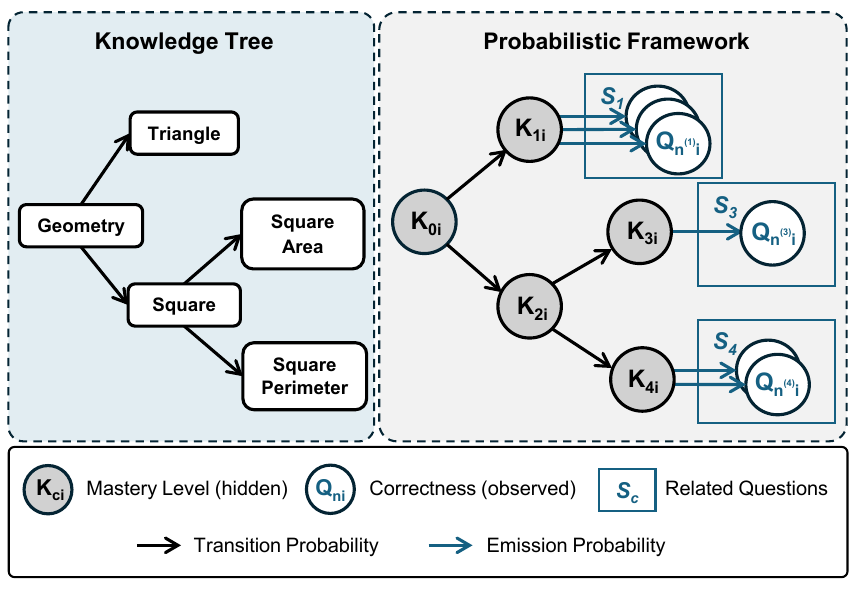}
  \caption{An KC tree structure (left), and its corresponding probabilistic framework (right). Student $i$'s knowledge state is represented by a set of hidden binary variables $K_{ci}$ over a tree-structure hierarchy. Each observed response $Q_{ni}$ is connected with a KC, and $\mathcal{S}_c$ denotes all associated responses of KC $c$.}
  \label{fig:assumption}
\end{figure}

\textbf{Third (Transition Probabilities)}, for each student $i$, the probability graphical model for $\{K_{ci}\}$ (Fig.~\ref{fig:assumption} (right)) follows the same topological structure as the knowledge concept tree (Fig.~\ref{fig:assumption} (left)). Namely, each $K_{ci}$ only directly depends on the mastery variable of its parent KC:
\begin{equation}
    p\big(\cup_{c} \{K_{ci}\}\big) = \prod_c p(K_{ci} | K_{\mathcal{P}(c)i}),
\end{equation}
where $\mathcal{P}(c)$ denotes the parent KC of $c$. To model each transition probability, {\small $p(K_{ci} | K_{\mathcal{P}(c)i})$}, we assume that mastering a parent KC would entail mastering ALL its children (but not vise versa), hence
\begin{equation}
p(K_{ci} = 1 \mid K_{\mathcal{P}(c)i} = k) = 
\begin{cases}
1 & \text{if } k = 1, \\
\gamma_c & \text{otherwise},
\end{cases}
\label{eq:transition}
\end{equation}
where each $\gamma_c$ is a parameter to be estimated. When $c$ is the root node, Eq.~\eqref{eq:transition} still applies with the condition {\small $K_{\mathcal{P}(c)i} = k$} removed.

\textbf{Fourth (Emission Probabilities)}, as shown in Fig.~\ref{fig:assumption} (right), we assume each $Q_{ni}$ only directly depends on the mastery variable of its corresponding labeled concepts, \emph{i.e.},
\begin{equation}
    p\big(\cup_{n} \{Q_{ni}\} | \cup_{c} \{K_{ci}\}\big) = \prod_n p(Q_{ni} | K_{\mathcal{M}(n)i}),
    \label{eq:emission_indep}
\end{equation}
where $\mathcal{M}(n)$ denotes the labeled KC for exercise $n$. Each emission probability, {\small $p(Q_{ni} | K_{\mathcal{M}(n)i})$}, is modeled as
\begin{equation}
    p(Q_{ni} = 1 \mid K_{\mathcal{M}(n)i} = k) =
    \begin{cases}
    \phi_n, & \text{if } k = 1, \\
    \varepsilon, & \text{otherwise,}
    \end{cases}
    \label{eq:emission}
\end{equation}
where $\phi_n$ represents the correct probability if the student knows the underlying concepts. It takes on three possible values, $\{r_{easy}, r_{med}, r_{hard}\}$, depending on the difficulty level of the exercise, which is determined by which of the three pre-defined bins (high, medium, or low) the historical solve rate falls into. $\varepsilon$ represents the correct probability by random guessing. $\varepsilon < r_{hard} < r_{med} < r_{easy}$ are parameters to be estimated.

\textbf{Summary.} The joint probability of all the hidden and observed variables is defined by Eqs.~\eqref{eq:indep} to \eqref{eq:emission}. The parameters of the model, $\bm \theta$, include
\begin{equation}
    \bm\theta = [\cup_c \gamma_c, r_{easy}, r_{med}, r_{hard}, \varepsilon].
\end{equation}

Sec.~\ref{subsec:estimation} will discuss how to estimate $\bm \theta$. Sec.~\ref{subsec:inference} will discuss how to predict the correctness (Eq.~\eqref{eq:pred}).

\subsection{Parameter Estimation}
\label{subsec:estimation}

The parameters $\bm \theta$ are estimated via the maximum log-likelihood objective on the \textit{observed variables}:
\begin{equation}
    \max_{\bm \theta} \log p_{\bm \theta}(\mathcal{Q}),
    \label{eq:mle}
\end{equation}
where $\mathcal{Q}$ denotes a set of observed historical correctness; {\small $p_{\bm \theta}(\mathcal{Q})$} can be computed by marginalizing Eq.~\eqref{eq:indep} over all the hidden variables, $\{K_{ci}\}$ (we add a subscript $\bm \theta$ to emphasize that $p$ is parameterized by $\bm \theta$).

Directly computing Eq.~\eqref{eq:mle} is computationally expensive due to the marginalization of the hidden variables. We thus adopt the standard EM algorithm \cite{EM} for the optimization. EM algorithm is an iterative algorithm. Denote $\bm \theta^{(\tau-1)}$ as the parameter estimate after iteration $\tau-1$. Then $\bm \theta^{(\tau)}$ can be computed from $\bm \theta^{(\tau-1)}$ via the following objective:
\begin{equation}
    \bm \theta^{(\tau)} = \argmax_{\bm \theta} A(\bm \theta; \bm \theta^{(\tau-1)}),
    \label{eq:em}
\end{equation}
where {\small $A(\bm \theta; \bm \theta^{(\tau-1)})$} equals

\begin{equation}
    \begin{aligned}
    \mathbb{E}_{p_{\bm \theta^{(\tau-1)}}(\cup_{c,i}\{K_{ci}\} | \mathcal{Q})} [\log p_{\bm \theta}(\cup_{n,i} \{K_{ci}\}, \mathcal{Q})].
    \end{aligned}
    \label{eq:likelihood}
\end{equation}

It can be shown that EM algorithm can converge to the optimal solution to Eq.~\eqref{eq:mle}. Eq.~\eqref{eq:em} bears a closed-form solution that can be computed efficiently, which is detailed in Appendix~\ref{appx:estimation}.

\subsection{Inference}
\label{subsec:inference}

Once $\bm \theta$ is estimated, we can predict if student $i$ can answer question $n^*$ correctly by computing Eq.~\eqref{eq:pred}, which can be further decomposed into
\begin{equation}
    \begin{aligned}
    & p_{\bm \theta}(Q_{n^*i} = 1 | \mathcal{Q}_i) \\
    = &\sum_{k=0}^ 1p_{\bm \theta} (K_{\mathcal{M}(n^*)i}=k | \mathcal{Q}_i) p(Q_{n^*i}=1 | K_{\mathcal{M}(n^*)i}=k, \mathcal{Q}_i) \\
    = & \sum_{k=0}^ 1 p_{\bm \theta} (K_{\mathcal{M}(n^*)i}=k | \mathcal{Q}_i) p(Q_{n^*i}=1 | K_{\mathcal{M}(n^*)i}=k) \\
    =&p_{\bm \theta} (K_{\mathcal{M}(n^*)i}=0 | \mathcal{Q}_i) \varepsilon + p_{\bm \theta} (K_{\mathcal{M}(n^*)i}=1 | \mathcal{Q}_i) \phi_n,
    \end{aligned}
    \label{eq:pred_decompose}
\end{equation}
where the second equality is derived from the conditional independence assumption in Eq.~\eqref{eq:emission_indep}; the last equality is from Eq.~\eqref{eq:emission}.

Eq.~\eqref{eq:pred_decompose} implies that the prediction process involves two steps. First, infer the mastery of the KC associated with the target exercise based on the historical performance, \emph{i.e.}, computing {\small $p_{\bm \theta} (K_{\mathcal{M}(n^*)i}=k | \mathcal{Q}_i)$}, which can be efficiently computed using the \emph{upward-downward algorithm} \cite{HMT} (see Appendix~\ref{appx:algorithm}). Second, use the KC mastery probability to modulate the emission probability (last line of Eq.~\eqref{eq:pred_decompose}).

\subsection{Incremental Update}

In online settings, each $\mathcal{Q}_i$ is constantly expanding, making it necessary to re-estimate $\bm \theta$ based on Eqs.~\eqref{eq:mle} and \eqref{eq:em}. We design the following incremental update scheme to balance between efficiency and performance, as shown in Fig.~\ref{fig:framework}.

\noindent \textbf{Communal Burn-in.} At the start, each $\mathcal{Q}_i$ contains only a few data. We aggregate the early data from all target students into a shared burn-in dataset, denoted by $\mathcal{Q}_{init}$, and estimate a common model, $\bm \theta_{init}$, via Eq.~\eqref{eq:mle} with $\mathcal{Q} = \mathcal{Q}_{init}$. We run the full EM iterations till convergence. This model is then used to predict the correctness of the first exercise (post-burn-in) for both seen and unseen students.

\noindent \textbf{Personalized Update.} After the burn-in phase, we maintain personalized models for each target student. Let $\bm \theta_i$ denote the model parameters for student $i$, estimated by Eq.~\eqref{eq:mle} with $\mathcal{Q} = \mathcal{Q}_{\text{init}} \cup \mathcal{Q}_i$ -- that is, the burn-in data from all students and all historical responses from student $i$. As new responses are observed (\emph{i.e.}, as $\mathcal{Q}_i$ expands), we incrementally update $\bm \theta_i$ by performing a single EM iteration, enabling efficient adaptation and personalization.

\section{Experiment}
We first describe the construction of classroom testbeds and baseline settings in Secs.~\ref{sec:data} and \ref{sec:setup}, respectively.
We then compare \algname with existing baselines and provide analysis in Sec.~\ref{sec:main} - Sec.~\ref{sec:ablation}.

\subsection{Data Construction} \label{sec:data}

We first construct evaluation datasets that follow the low-resource scenarios as described in Sec.~\ref{sec:intro} and that come with knowledge concept trees. We will open-source all our datasets, along with the code for preprocessing and dataset construction.

\paragraph{Dataset Selection.} We identify two widely used educational datasets: \textsc{XES3G5M}~\cite{XES3G5M} for K–12 math and \textsc{MOOCRadar}~\cite{MOOCRadar} for university-level courses. Each exercise is annotated with fine-grained KCs and can also be assigned a difficulty label (easy/medium/hard) based on student correctness rates. Statistics for the datasets are in Appendix~\ref{appx:data-stats}.

\paragraph{Knowledge Concept Tree.} 
Next, we discuss how the knowledge concept trees are structured in these datasets.
\textsc{XES3G5M} arranges its KCs into a hierarchical entailment structure, allowing us to use it directly as a knowledge concept tree.
For \textsc{MOOCRadar}, it only provides KCs without any hierarchical structure, so we build a knowledge concept tree for it instead.
Inspired by KCQRL~\cite{KCQRL}, we encode KCs into semantic embeddings~\cite{mcinnes2018umap} and cluster semantically similar KCs together~\cite{campello2013density}. For each cluster, we use GPT-4o-mini~\cite{Gpt4omini} to generate a representative KC label. Finally, we manually annotate unclustered KCs. Further details are provided in Appendix~\ref{appx:kc-tree-mooc}.

\paragraph{Data Sampling.} To simulate the low-resource setting, we need to sample a subset of questions and students. First, to sample a subset of questions and knowledge concepts, we partition the knowledge concept trees by taking each level-1 node as the root of a knowledge module. Each module consists of the root and all its descendant KCs. We then identify students who have worked on at least 50 exercises associated with the module. For each dataset, we select the top-3 knowledge modules with the largest number of students satisfying the condition. Then, we remove out-of-module exercises for those students.

\begin{table}[h]
\small
\centering
\setlength{\tabcolsep}{5pt}
\begin{tabular}{lccc}
\toprule
\rowcolor[HTML]{E6E6E6}
\textbf{\textsc{XES3G5M}} & \textbf{Mean} & \textbf{Std} & \textbf{Interactions} \\
\midrule
Application Module & 0.6249 & 0.1600 & 10532 \\
Computation Module   & 0.6235 & 0.1354 & 7521 \\
Counting Module   & 0.6689 & 0.1424 & 7279 \\
\toprule
\rowcolor[HTML]{E6E6E6}
\textbf{\textsc{MOOCRadar}} & \textbf{Mean} & \textbf{Std} & \textbf{Interactions} \\
\midrule
Wine Knowledge   & 0.6592 & 0.1342 & 8469 \\
Circuit Design   & 0.6394 & 0.1320 & 17616 \\
Education Theory & 0.6791 & 0.1205 & 8796 \\
\hline
\end{tabular}
\caption{Statistics of simulated classroom data extracted from \textsc{XES3G5M} and \textsc{MOOCRadar}. Mean and Std correspond to class's accuracy rates on all interactions.}
\label{tab:classroom}
\vspace{-3mm}
\end{table}

We then sample 100 students per module with a good coverage of overall performance. Specifically, we use a normal distribution $\mathcal{N}(0.65,0.15)$ to draw 100 overall correctness rates. For each drawn correctness rate, we select a student whose overall correctness rate best matches the drawn sample without replacement. Each sampled student's first 10 interactions are added to the burn-in data, and the remaining are used for incremental inference. The statistics of the classroom sets are shown in Table~\ref{tab:classroom}.

\subsection{Setup} \label{sec:setup}

\paragraph{Metrics.} Following previous works, we evaluate models using standard binary classification metrics, including AUC, accuracy, and F1 score.

\paragraph{Baselines.} We consider several commonly used methods in knowledge tracing, including conventional deep learning knowledge tracing (DLKT) methods and LLM-based approaches.

$\bullet$ \textsc{AKT}~\cite{AKT} is a transformer-based KT model that uses a monotonic attention mechanism to measure the relevance between current questions and historical exercises.

$\bullet$ \textsc{SAINT}~\cite{SAINT} is a transformer-based model with a deep encoder-decoder architecture. It separates exercise and response sequences and encodes them respectively, allowing stacked attention layers to learn the dependencies.
    
$\bullet$ \textsc{qDKT}~\cite{qDKT} is a variant of DKT that models performance at the question level. It applies Laplacian regularization to smooth predictions across similar questions and uses fastText based initialization to improve generalization.
    
$\bullet$ \textsc{LLM} has demonstrated remarkable performance across a wide range of tasks. Their ability to handle long-context inputs makes them potentially capable for modeling student competence based on historical exercises and predicting future performance. We adopt two different LLM architectures: \textsc{Qwen-2.5-7B}~\cite{bai2025qwen2d5} and \textsc{LLaMA-3.2-3B}~\cite{llama3d2}. During inference, each model is given 10 historical exercises of the current student and predicts the correctness of the next response.

To ensure a fair comparison with our incremental framework, we implement an online variant for each DLKT baseline, denoted as \textsc{AKT-Online}, \textsc{SAINT-Online}, and \textsc{qDKT-Online}. These models are initially trained on the same burn-in data as their offline versions. During inference, we aggregate all students' cumulative interactions into the training set and perform an additional round of training. The updated model is then used to predict the next question for all students. Note that this online strategy gives the baselines access to more training data compared to \algname, which only performs individual-level updates without using other students’ new interactions.

For all DLKT baselines above, we adopt the standardized implementation provided by \textsc{pyKT}~\cite{pyKT}, a unified Python library for benchmarking KT models. For all LLM baselines, we adopt the publicly released checkpoints.

{\renewcommand{\arraystretch}{1.4}
\begin{table*}[t]
  \small
  \centering
  \setlength{\tabcolsep}{4.2pt}
  \resizebox{\textwidth}{!}{
  \begin{tabular}{l|ccc!{\vrule width 1pt}ccc!{\vrule width 1pt}ccc!{\vrule width 1pt}ccc}
    \toprule
    \textbf{Model} & \textbf{AUC}& \textbf{ACC} & \textbf{F1} & \textbf{AUC}& \textbf{ACC} & \textbf{F1} & \textbf{AUC}& \textbf{ACC} & \textbf{F1} & \textbf{AUC}& \textbf{ACC} & \textbf{F1} \\ 
    \midrule
    \rowcolor[HTML]{D9E3F0} \multicolumn{13}{c}{\textbf{\textsc{XES3G5M}}~\cite{XES3G5M}} \\
    \hline
    & \multicolumn{3}{c!{\vrule width 1pt}}{\textbf{Application Module}} 
    & \multicolumn{3}{c!{\vrule width 1pt}}{\textbf{Computation Module}}
    & \multicolumn{3}{c!{\vrule width 1pt}}{\textbf{Counting Module}}
    & \multicolumn{3}{c}{\textbf{Avg.}}\\
    \hline
    \textsc{AKT}~\cite{AKT}           & 0.6790 & 0.6659 & 0.7418 & 0.6701 & 0.6628 & 0.7560 & 0.6994 & 0.7184 & 0.8114 & 0.6828 & 0.6824 & 0.7697 \\
    \textsc{AKT-Online}    & \underline{0.7106} & 0.6825 & 0.7645 & \underline{0.6848} & \underline{0.6722} & 0.7690 & \underline{0.7174} & \underline{0.7203} & 0.8080 & \underline{0.7043} & \underline{0.6917} & 0.7805 \\
    \textsc{SAINT}~\cite{SAINT}         & 0.6031 & 0.6423 & 0.7822 & 0.6079 & 0.6306 & \underline{0.7735} & 0.5794 & 0.6886 & \underline{0.8156} & 0.5630 & 0.6608 & \underline{0.7956} \\
    \textsc{SAINT-Online}  & 0.6899 & \underline{0.6826} & 0.7615 & 0.6641 & 0.6609 & 0.7558 & 0.6779 & 0.6904 & 0.7834 & 0.6773 & 0.6780 & 0.7669  \\
    \textsc{qDKT}~\cite{qDKT}          & 0.5177 & 0.5178 & 0.5965 & 0.5338 & 0.5272 & 0.5897 & 0.4856 & 0.4716 & 0.5410 & 0.5124 & 0.5055 & 0.5757  \\
    \textsc{qDKT-Online}   & 0.6663 & 0.6358 & 0.7185 & 0.6606 & 0.6363 & 0.7138 & 0.6988 & 0.6697 & 0.7603 & 0.6752 & 0.6473 & 0.7309  \\
    \textsc{Qwen-2.5}~\cite{bai2025qwen2d5}        & 0.5939 & 0.6425 & 0.7823 & 0.5730 &  0.6301 & 0.7731 & 0.5918 & 0.6857 & 0.8136 & 0.5862    &  0.6528   & 0.7897    \\
    \textsc{Llama-3.2}~\cite{llama3d2}        & 0.5839 & 0.6431 & \underline{0.7825} & 0.5649 & 0.6304 & 0.7732 & 0.5873 &  0.6868 & 0.8136 & 0.5787    &  0.6534   & 0.7898    \\
    \rowcolor{gray!20}
    \textbf{\textsc{Ours}} & \textbf{0.7448} & \textbf{0.7057} & \textbf{0.7962} & \textbf{0.7079} & \textbf{0.6952} & \textbf{0.7807} & \textbf{0.7470} & \textbf{0.7326} & \textbf{0.8258} & \textbf{0.7332} & \textbf{0.7111} & \textbf{0.8009} \\
    \midrule
    \rowcolor[HTML]{D9E3F0} \multicolumn{13}{c}{\textbf{\textsc{MoocRadar}}~\cite{MOOCRadar}} \\
    \hline 
    & \multicolumn{3}{c!{\vrule width 1pt}}{\textbf{Wine Knowledge}} 
    & \multicolumn{3}{c!{\vrule width 1pt}}{\textbf{Circuit Design}}
    & \multicolumn{3}{c!{\vrule width 1pt}}{\textbf{Education Theory}}
    & \multicolumn{3}{c}{\textbf{Avg.}}\\
    \hline
    \textsc{AKT}~\cite{AKT}           & 0.5901 & 0.6546 & 0.7784 & 0.6142 & 0.6402 & 0.7722 & 0.5535 & 0.6910 & \underline{0.8173} & 0.5859 & 0.6619 & 0.7893  \\
    \textsc{AKT-Online}    & 0.7208 & 0.6967 & \underline{0.7955} & 0.6726 & 0.6712 & 0.7749 & 0.6442 & 0.6974 & 0.8059  & 0.6792 & 0.6884 & 0.7921 \\
    \textsc{SAINT}~\cite{SAINT}         & 0.4905 & 0.6568 & 0.7929 & 0.6066 & 0.6347 & 0.7766 & 0.5918 & 0.6910 & \underline{0.8173} & 0.5630 & 0.6608 & \underline{0.7956} \\
    \textsc{SAINT-Online}  & 0.5646 & 0.6510 & 0.7767 & 0.6284 & 0.6389 & 0.7554 & 0.6052 & 0.6878 & 0.8085 & 0.5994 & 0.6592 & 0.7802 \\
    \textsc{qDKT}~\cite{qDKT}          & 0.5857 & 0.5629 & 0.6371 & 0.5248 & 0.5106 & 0.5774 & 0.5420 & 0.5142 & 0.5918 & 0.5508 & 0.5292 & 0.6021 \\
    \textsc{qDKT-Online}   & \underline{0.7661} & \underline{0.7230} & 0.7924 & \underline{0.7063} & \underline{0.6716} & 0.7438 & \underline{0.7327} & \underline{0.7292} & 0.8118 & \underline{0.7350} & \underline{0.7079} & 0.7827  \\
    \textsc{Qwen-2.5}~\cite{bai2025qwen2d5}        & 0.6190 & 0.6526 & 0.7898 & 0.5595 & 0.6377 & 0.7788 & 0.5454 & 0.6861 & 0.8138 & 0.5746    & 0.6588    &  0.7941   \\
    \textsc{Llama-3.2}~\cite{llama3d2}        & 0.5889 & 0.6526 & 0.7898 & 0.5852 & 0.6385 & \underline{0.7790} & 0.5562 & 0.6887 & 0.8142 & 0.5768    &  0.6599   & 0.7943    \\
    \rowcolor{gray!20}
    \textbf{\textsc{Ours}} & \textbf{0.7714} & \textbf{0.7354} & \textbf{0.8237} & \textbf{0.7662} & \textbf{0.7315} & \textbf{0.7906} & \textbf{0.7891} & \textbf{0.7656} & \textbf{0.8329} & \textbf{0.7756} & \textbf{0.7442} & \textbf{0.8157}  \\
    \midrule
  \end{tabular}
  }
  \vspace{-2mm}
  \caption{Knowledge tracing performance comparison on \textsc{XES3G5M} and \textsc{MoocRadar}. \textbf{Bold} number indicates the best performance in each module, and \underline{underlined} number indicates the second best. The Avg. column reports the average result across all modules in each dataset. }
  \label{tab:main-results}
  \vspace{-2mm}
\end{table*}
}

\subsection{Main Result}\label{sec:main}
Table~\ref{tab:main-results} presents the performance of all methods across both datasets. We summarize our key observations as follows:

First, \algname achieves the highest scores across all metrics on both \textsc{XES3G5M} and \textsc{MOOCRadar}, demonstrating its effectiveness and robustness.
Second, among the conventional DLKT baselines, all online variants outperform their offline counterparts, highlighting the importance of online updating.
Notably, compared with our method, all DLKT-Online baselines have access to more data at each update step, as they retrain using the full classroom’s cumulative interactions. In contrast, \algname updates only with the new interaction from a single student.
Despite this apparent advantage, these baselines still underperform, suggesting that simply applying online retraining to existing DLKT models is insufficient to fully capture the evolution of an individual student's knowledge state.
Third, we observe that both Qwen and LLaMA do not perform well on both datasets. This suggests that current LLMs may lack the capability to effectively extract student characteristics from historical data and make accurate predictions.

\subsection{Qualitative Analysis}
To better understand how our framework performs real-time updates, we visualize the changes in the posterior mastery probability, {\small $p_{\bm \theta} (K_{ci}=1 | \mathcal{Q}_i)$}, across a KC subtree during incremental inference for two students (Fig.~\ref{fig:tree}), as the historical observation, {\small $\mathcal{Q}_i$}, expands. Color reflects differences between the current mastery probability and the initial estimation based on burn-in data only. Each node is annotated with step-wise mastery change. Our findings are as follows:

\begin{figure*}[t]
  \centering
  \includegraphics[width=\textwidth]{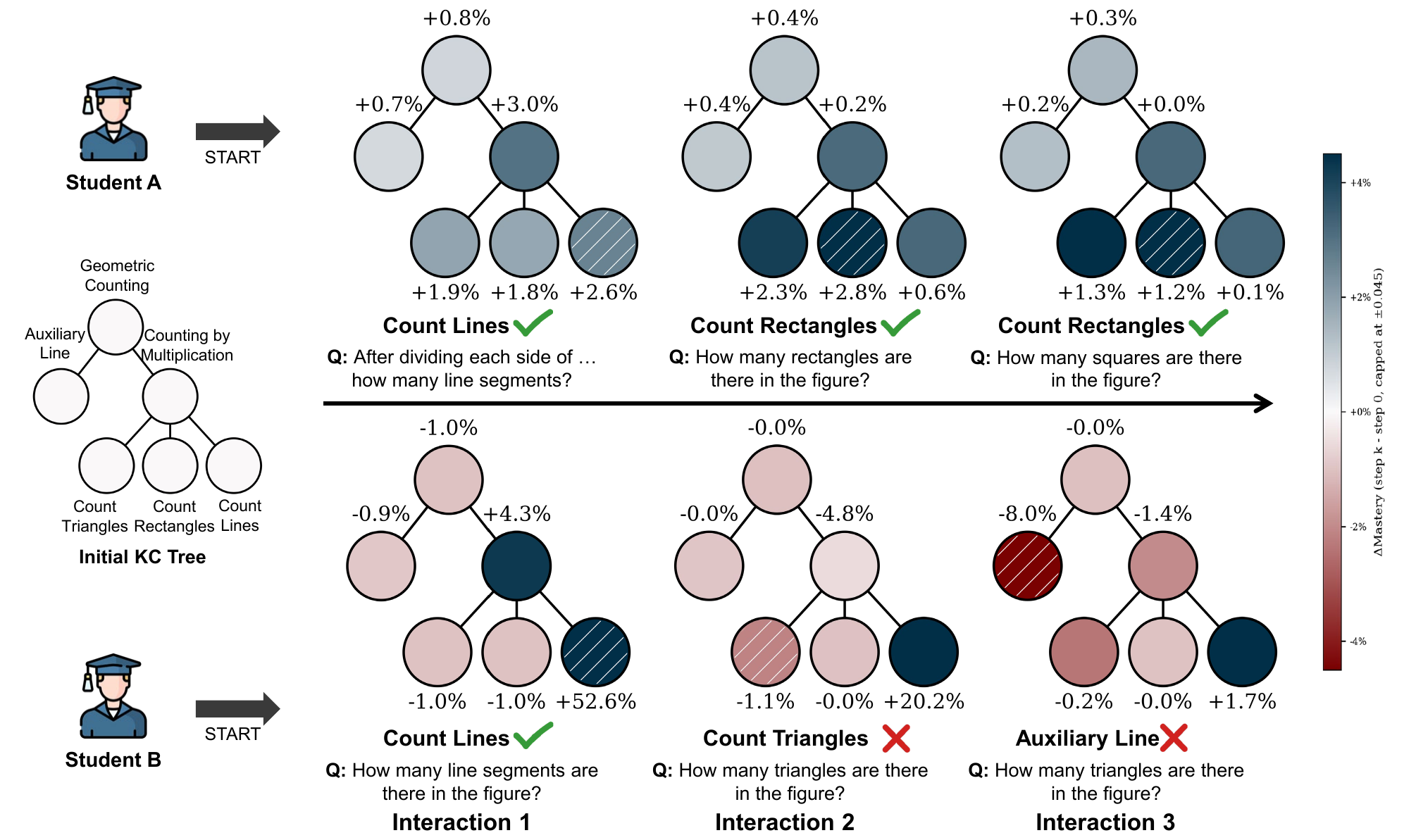}
  \vspace{-6mm}
  \caption{Examples of KC mastery probability update.
  The node color indicates the cumulative mastery probability change from the initial estimation (after burn-in) to the current step, with blue for increase and red for decrease. The value inside denotes the step-wise change from the previous step. The node with dashed lines indicates the concept associated with the current question.
  The bottom row of each tree illustrates the student's interaction at that step.
  }
  \vspace{-2mm}
  \label{fig:tree}
\end{figure*}

First, we find the probability updates align with the inputs. In this example, Student A solves all three questions correctly, leading to an increase in mastery probability on all nodes in the tree. Student B answers the first question correctly, but the next two incorrectly. As a result, only the ``Count Line'' node corresponding to the first question shows a higher mastery probability than the initial estimation, while the rest of the nodes all get lower mastery probabilities.

Second, it is worth noting that due to the tree propagation, the mastery probability changes are not confined to the KCs directly associated with the questions. Other nodes also dynamically adjust their mastery probabilities, even when the KC has not been explicitly visited in the student's exercise history. This effect is evident not only in direct parent or sibling nodes. For example, in Student A's first interaction, correctly answering a ``Count Lines'' question also increases confidence in related ``Auxiliary Line'' nodes.

In summary, these findings demonstrate the ability of \algname to dynamically update and adjust the student's profile in a structured and interpretable way, with respect to the concept hierarchy.

\subsection{Ablation Study}\label{sec:ablation}
\begin{figure}[t]
  \centering
  \includegraphics[width=0.48\textwidth]{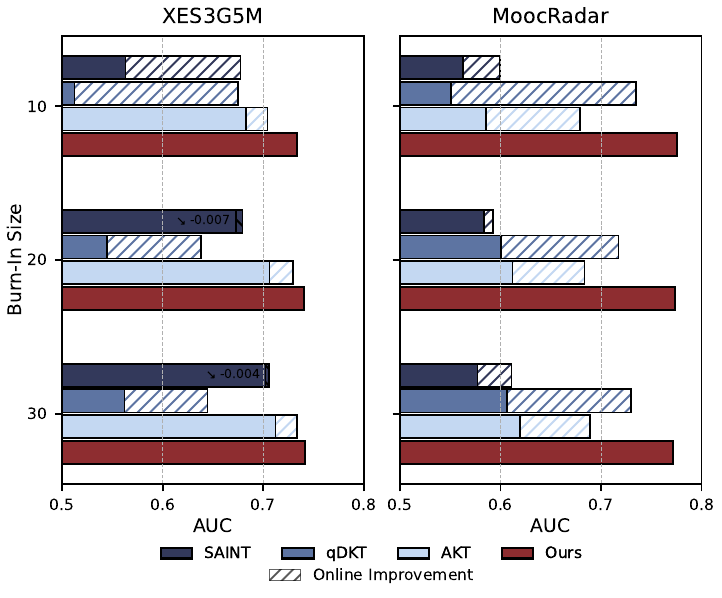}
  \vspace{-4mm}
  \caption{Effect of burn-in size. We vary the number of burn-in exercises per student and compare the performance of models. Dashed lines represent online improvement; negative gains are annotated inline.}
  \vspace{-3mm}
  \label{fig:Ablation_AUC}
\end{figure}

We conduct an ablation study to investigate how the amount of available burn-in data affects model performance. Fig.~\ref{fig:Ablation_AUC} shows the AUC results on both datasets (average across all modules) under different burn-in sizes. We exclude LLM-based KT methods from this comparison, as their few-shot prompting approach relies on randomly selected examples rather than directly leveraging the structured burn-in set used by other methods.

As observed, our model consistently outperforms the baselines across all burn-in size settings, particularly in low burn-in scenarios. As the burn-in size increases, all offline methods show improved performance, and as expected, they benefit less from online updates. 
Nevertheless, \algname remains competitive even under larger burn-in sizes.
The fact that DLKT models require a substantial amount of burn-in data to perform well further underscores their limitations in real-world settings, where data is often scarce and the ability to understand students' capabilities from a limited history of exercises is crucial.

\section{Related Work}

\noindent \textbf{Deep Learning KT}
In recent years, with the rapid development of deep learning techniques, many knowledge tracing methods have adopted neural models to predict student performance. Early KT models include RNN-based DKT~\cite{DKT}, memory-augmented DKVMN~\cite{DKVMN}, and its IRT~\cite{IRT}-inspired neural variant Deep-IRT~\cite{Deep-IRT}. Later works such as SAKT~\cite{SAKT} and GIKT~\cite{GIKT} adopt attention mechanisms to better model the importance of historical exercises. Transformer-based KT models, such as AKT~\cite{AKT}, qDKT~\cite{qDKT}, SAINT~\cite{SAINT}, and AT-DKT~\cite{AT-DKT}, further improve this by capturing long-range dependencies and contextual signals. These models typically treat KCs as independent and represent the student knowledge in latent vectors, limiting interpretability and generalizability.

\noindent \textbf{Structure-Aware KT} Several works have attempted to incorporate structural relationships among KCs. Recent graph-based KT models such as GKT~\cite{GKT}, GIKT~\cite{GIKT}, SHDKT~\cite{SHDKT}, KSGAN~\cite{KSGAN}, and DHKT~\cite{DHKT} enhance embeddings by encoding concept co-occurrence or hierarchy via GNNs or attention. More recent approaches like HHSKT~\cite{HHSKT} construct heterogeneous graphs over learner-question interactions with hierarchical differentiation, while PSI-KT~\cite{PSIKT} jointly models learner traits and prerequisite structure using a hierarchical Bayesian generative model. These models primarily treat structure as auxiliary input or require substantial learner histories for inference.

\noindent \textbf{LLM-Based KT} LLMs have seen remarkable progress in recent years~\cite{GPT,Claude,Gemini}, leading to growing interest in their application in education, including online tutoring, feedback generation, and question generation~\cite{heffernan2024leveraging, liu2024scaffolding, luo2024chain}. Recent works have explored applying LLMs for KT, aiming to improve interpretability and reduce data dependence. One line of work adopts few-shot prompting to estimate student mastery and generate natural language rationales~\cite{FSP}. These methods require repeated prompting, are sensitive to example selection, and lead to high computational and financial costs. Another line of work investigates fine-tuning LLMs for KT. Current results show that fine-tuned LLMs can match classical probabilistic models like BKT~\cite{BKT}, but they still lag behind modern deep KT models' performance~\cite{FT-LLM, LKT}. These studies suggest that while LLMs offer flexibility and reasoning capabilities, further adaptation is needed for real-world deployment.

\section{Conclusion}\label{sec:conclusion}
In this paper, we propose \algname, a probabilistic framework for low-resource, online knowledge tracing. Leveraging the hierarchical structure of KCs, \algname builds a Hidden Markov Tree Model, where hidden variables represent concept mastery levels and observed variables correspond to students’ correctness on individual exercise questions. 
It enables effective learning from minimal initial data and supports incremental updates as new responses are observed. Experiments across six simulated classroom scenarios show that \algname consistently outperforms existing baselines, including online variants of conventional deep learning models and LLM-based methods. Qualitative analysis further demonstrates the ability of \algname to dynamically refine mastery estimates and propagate the updates across related concepts.
Overall, our work underscores the value of integrating concept hierarchies and probabilistic inference in developing practical, personalized knowledge tracing systems.

\section{Limitations}

Our current framework relies on several simplifying assumptions to ensure tractable inference and model interpretability. We assume that each exercise is associated with only a single KC, and each KC only has one parent. These assumptions ensure the valid tree structure holds and efficient tree-based propagation, but may limit the applicability to exercises involving multiple concepts, where the information of extra KCs are not taken into consideration in our framework.

Besides, while \algname generalizes well to unseen students through probabilistic inference and incremental updates, it currently utilizes only binary correctness labels to estimate student performance and does not support richer interaction types, such as open-ended responses or partial credits.

\section{Acknowledgement}
Xinyi Gao, Qiucheng Wu, Xuechen Liu, Xu Ying, and Shiyu Chang acknowledge support from the National Science Foundation (NSF) under Grant IIS-2302730. Additionally, Xinyi Gao, Qiucheng Wu, and Shiyu Chang acknowledge support from the Accelerate Foundation Models Research (AFMR) Program of Microsoft.

\bibliographystyle{unsrt}
\bibliography{reference}

\begin{thebibliography}{10}

\bibitem{DKT}
C~Piech, J~Bassen, J~Huang, S~Ganguli, M~Sahami, LJ~Guibas, and J~Sohl-Dickstein.
\newblock Deep knowledge tracing. advances in neural information processing systems.
\newblock {\em Association for Computing Machinery}, pages 201--204, 2015.

\bibitem{AKT}
Aritra Ghosh, Neil Heffernan, and Andrew~S Lan.
\newblock Context-aware attentive knowledge tracing.
\newblock In {\em Proceedings of the 26th ACM SIGKDD international conference on knowledge discovery \& data mining}, pages 2330--2339, 2020.

\bibitem{DKVMN}
Jiani Zhang, Xingjian Shi, Irwin King, and Dit-Yan Yeung.
\newblock Dynamic key-value memory networks for knowledge tracing.
\newblock In {\em Proceedings of the 26th international conference on World Wide Web}, pages 765--774, 2017.

\bibitem{SAKT}
Shalini Pandey and George Karypis.
\newblock A self-attentive model for knowledge tracing.
\newblock {\em arXiv preprint arXiv:1907.06837}, 2019.

\bibitem{simpleKT}
Zitao Liu, Qiongqiong Liu, Jiahao Chen, Shuyan Huang, and Weiqi Luo.
\newblock simplekt: a simple but tough-to-beat baseline for knowledge tracing.
\newblock {\em arXiv preprint arXiv:2302.06881}, 2023.

\bibitem{FSP}
Haoxuan Li, Jifan Yu, Yuanxin Ouyang, Zhuang Liu, Wenge Rong, Juanzi Li, and Zhang Xiong.
\newblock Explainable few-shot knowledge tracing.
\newblock {\em arXiv preprint arXiv:2405.14391}, 2024.

\bibitem{FT-LLM}
Seyed~Parsa Neshaei, Richard~Lee Davis, Adam Hazimeh, Bojan Lazarevski, Pierre Dillenbourg, and Tanja K{\"a}ser.
\newblock Towards modeling learner performance with large language models.
\newblock {\em arXiv preprint arXiv:2403.14661}, 2024.

\bibitem{HMT}
Matthew~S Crouse, Robert~D Nowak, and Richard~G Baraniuk.
\newblock Wavelet-based statistical signal processing using hidden markov models.
\newblock {\em IEEE Transactions on signal processing}, 46(4):886--902, 2002.

\bibitem{EM}
Arthur~P Dempster, Nan~M Laird, and Donald~B Rubin.
\newblock Maximum likelihood from incomplete data via the em algorithm.
\newblock {\em Journal of the royal statistical society: series B (methodological)}, 39(1):1--22, 1977.

\bibitem{XES3G5M}
Zitao Liu, Qiongqiong Liu, Teng Guo, Jiahao Chen, Shuyan Huang, Xiangyu Zhao, Jiliang Tang, Weiqi Luo, and Jian Weng.
\newblock Xes3g5m: A knowledge tracing benchmark dataset with auxiliary information.
\newblock {\em Advances in Neural Information Processing Systems}, 36:32958--32970, 2023.

\bibitem{MOOCRadar}
Jifan Yu, Mengying Lu, Qingyang Zhong, Zijun Yao, Shangqing Tu, Zhengshan Liao, Xiaoya Li, Manli Li, Lei Hou, Hai-Tao Zheng, et~al.
\newblock Moocradar: A fine-grained and multi-aspect knowledge repository for improving cognitive student modeling in moocs.
\newblock In {\em Proceedings of the 46th International ACM SIGIR Conference on Research and Development in Information Retrieval}, pages 2924--2934, 2023.

\bibitem{KCQRL}
Yilmazcan Ozyurt, Stefan Feuerriegel, and Mrinmaya Sachan.
\newblock Automated knowledge concept annotation and question representation learning for knowledge tracing.
\newblock {\em arXiv preprint arXiv:2410.01727}, 2024.

\bibitem{mcinnes2018umap}
Leland McInnes, John Healy, and James Melville.
\newblock Umap: Uniform manifold approximation and projection for dimension reduction.
\newblock {\em arXiv preprint arXiv:1802.03426}, 2018.

\bibitem{campello2013density}
Ricardo~JGB Campello, Davoud Moulavi, and J{\"o}rg Sander.
\newblock Density-based clustering based on hierarchical density estimates.
\newblock In {\em Pacific-Asia conference on knowledge discovery and data mining}, pages 160--172. Springer, 2013.

\bibitem{Gpt4omini}
OpenAI.
\newblock Gpt-4o mini: advancing cost-efficient intelligence.
\newblock https://openai.com/index/gpt-4o-mini-advancing-cost-efficient-intelligence/, 2024.

\bibitem{SAINT}
Youngduck Choi, Youngnam Lee, Junghyun Cho, Jineon Baek, Byungsoo Kim, Yeongmin Cha, Dongmin Shin, Chan Bae, and Jaewe Heo.
\newblock Towards an appropriate query, key, and value computation for knowledge tracing.
\newblock In {\em Proceedings of the seventh ACM conference on learning@ scale}, pages 341--344, 2020.

\bibitem{qDKT}
Shashank Sonkar, Andrew~E Waters, Andrew~S Lan, Phillip~J Grimaldi, and Richard~G Baraniuk.
\newblock qdkt: Question-centric deep knowledge tracing.
\newblock {\em arXiv preprint arXiv:2005.12442}, 2020.

\bibitem{bai2025qwen2d5}
Qwen-Team.
\newblock Qwen2. 5 technical report.
\newblock {\em arXiv preprint arXiv:2502.13923}, 2025.

\bibitem{llama3d2}
Meta.
\newblock Llama 3.2: Revolutionizing edge ai and vision with open, customizable models.
\newblock https://ai.meta.com/blog/llama-3-2-connect-2024-vision-edge-mobile-devices/, 2024.

\bibitem{pyKT}
Zitao Liu, Qiongqiong Liu, Jiahao Chen, Shuyan Huang, Jiliang Tang, and Weiqi Luo.
\newblock pykt: a python library to benchmark deep learning based knowledge tracing models.
\newblock {\em Advances in Neural Information Processing Systems}, 35:18542--18555, 2022.

\bibitem{IRT}
Li~Cai, Kilchan Choi, Mark Hansen, and Lauren Harrell.
\newblock Item response theory.
\newblock {\em Annual Review of Statistics and Its Application}, 3(1):297--321, 2016.

\bibitem{Deep-IRT}
Chun-Kit Yeung.
\newblock Deep-irt: Make deep learning based knowledge tracing explainable using item response theory.
\newblock {\em arXiv preprint arXiv:1904.11738}, 2019.

\bibitem{GIKT}
Yang Yang, Jian Shen, Yanru Qu, Yunfei Liu, Kerong Wang, Yaoming Zhu, Weinan Zhang, and Yong Yu.
\newblock Gikt: a graph-based interaction model for knowledge tracing.
\newblock In {\em Machine learning and knowledge discovery in databases: European conference, ECML PKDD 2020, Ghent, Belgium, September 14--18, 2020, proceedings, part I}, pages 299--315. Springer, 2021.

\bibitem{AT-DKT}
Zitao Liu, Qiongqiong Liu, Jiahao Chen, Shuyan Huang, Boyu Gao, Weiqi Luo, and Jian Weng.
\newblock Enhancing deep knowledge tracing with auxiliary tasks.
\newblock In {\em Proceedings of the ACM web conference 2023}, pages 4178--4187, 2023.

\bibitem{GKT}
Hiromi Nakagawa, Yusuke Iwasawa, and Yutaka Matsuo.
\newblock Graph-based knowledge tracing: modeling student proficiency using graph neural network.
\newblock In {\em IEEE/WIC/aCM international conference on web intelligence}, pages 156--163, 2019.

\bibitem{SHDKT}
Zhenyuan Yang, Shimeng Xu, Changbo Wang, and Gaoqi He.
\newblock Skill-oriented hierarchical structure for deep knowledge tracing.
\newblock In {\em 2022 IEEE 34th International Conference on Tools with Artificial Intelligence (ICTAI)}, pages 425--432. IEEE, 2022.

\bibitem{KSGAN}
Shun Mao, Jieyu Zhan, Jiawei Li, and Yuncheng Jiang.
\newblock Knowledge structure-aware graph-attention networks for knowledge tracing.
\newblock In {\em International Conference on Knowledge Science, Engineering and Management}, pages 309--321. Springer, 2022.

\bibitem{DHKT}
Tianqi Wang, Fenglong Ma, and Jing Gao.
\newblock Deep hierarchical knowledge tracing.
\newblock In {\em Proceedings of the 12th international conference on educational data mining}, 2019.

\bibitem{HHSKT}
Qin Ni, Tingjiang Wei, Jiabao Zhao, Liang He, and Chanjin Zheng.
\newblock Hhskt: A learner--question interactions based heterogeneous graph neural network model for knowledge tracing.
\newblock {\em Expert Systems with Applications}, 215:119334, 2023.

\bibitem{PSIKT}
Hanqi Zhou, Robert Bamler, Charley~M Wu, and {\'A}lvaro Tejero-Cantero.
\newblock Predictive, scalable and interpretable knowledge tracing on structured domains.
\newblock In {\em The Twelfth International Conference on Learning Representations}, 2024.

\bibitem{GPT}
Josh Achiam, Steven Adler, Sandhini Agarwal, Lama Ahmad, Ilge Akkaya, Florencia~Leoni Aleman, Diogo Almeida, Janko Altenschmidt, Sam Altman, Shyamal Anadkat, et~al.
\newblock Gpt-4 technical report.
\newblock {\em arXiv preprint arXiv:2303.08774}, 2023.

\bibitem{Claude}
Anthropic.
\newblock Claude 3.7 sonnet and claude code.
\newblock https://www.anthropic.com/news/claude-3-7-sonnet, 2025.

\bibitem{Gemini}
Google DeepMind.
\newblock Gemini: Google deepmind's multimodal llms.
\newblock https://deepmind.google/technologies/gemini/, 2025.

\bibitem{heffernan2024leveraging}
Neil Heffernan, Rose Wang, Christopher MacLellan, Arto Hellas, Chenglu Li, Candace Walkington, Joshua Littenberg-Tobias, David Joyner, Steven Moore, Adish Singla, et~al.
\newblock Leveraging large language models for next-generation educational technologies.
\newblock In {\em Proceedings of the 17th International Conference on Educational Data Mining}, pages 1037--1039, 2024.

\bibitem{liu2024scaffolding}
Zhengyuan Liu, Stella~Xin Yin, Carolyn Lee, and Nancy~F Chen.
\newblock Scaffolding language learning via multi-modal tutoring systems with pedagogical instructions.
\newblock In {\em 2024 IEEE Conference on Artificial Intelligence (CAI)}, pages 1258--1265. IEEE, 2024.

\bibitem{luo2024chain}
Haohao Luo, Yang Deng, Ying Shen, See-Kiong Ng, and Tat-Seng Chua.
\newblock Chain-of-exemplar: enhancing distractor generation for multimodal educational question generation.
\newblock ACL, 2024.

\bibitem{BKT}
Albert~T Corbett and John~R Anderson.
\newblock Knowledge tracing: Modeling the acquisition of procedural knowledge.
\newblock {\em User modeling and user-adapted interaction}, 4:253--278, 1994.

\bibitem{LKT}
Unggi Lee, Jiyeong Bae, Dohee Kim, Sookbun Lee, Jaekwon Park, Taekyung Ahn, Gunho Lee, Damji Stratton, and Hyeoncheol Kim.
\newblock Language model can do knowledge tracing: Simple but effective method to integrate language model and knowledge tracing task.
\newblock {\em arXiv preprint arXiv:2406.02893}, 2024.

\bibitem{bge_embedding}
Shitao Xiao, Zheng Liu, Peitian Zhang, and Niklas Muennighoff.
\newblock C-pack: Packaged resources to advance general chinese embedding, 2023.

\bibitem{kwon2023efficient}
Woosuk Kwon, Zhuohan Li, Siyuan Zhuang, Ying Sheng, Lianmin Zheng, Cody~Hao Yu, Joseph~E. Gonzalez, Hao Zhang, and Ion Stoica.
\newblock Efficient memory management for large language model serving with pagedattention.
\newblock In {\em Proceedings of the ACM SIGOPS 29th Symposium on Operating Systems Principles}, 2023.

\end{thebibliography}

\appendix

\clearpage
\section{Parameter Estimation} \label{appx:estimation}

In this section, we provide the closed-form solution of $\bm \theta^{(\tau)}$, computed by taking the partial derivative on Eq.~\eqref{eq:likelihood} with respect to each parameter.
\begin{gather}
\gamma_{ci}^{\text{master } \mathcal{P}(c)} = p_{\bm \theta^{(\tau-1)}}(K_{ci} = 1, K_{\mathcal{P}(c)i} = 0|\mathcal{Q}_i) \\
\gamma_{ci}^{\text{not-master } \mathcal{P}(c)} = p_{\bm \theta^{(\tau-1)}}(K_{ci} = 0, K_{\mathcal{P}(c)i} = 0|\mathcal{Q}_i) \\
 \gamma_c^{(\tau)} = \frac{\sum_i \gamma_{ci}^{(1)}}{\sum_i \gamma_{ci}^{(0)} + \sum_i \gamma_{ci}^{(1)}}.
\end{gather}

The closed-form solution of any root node can be obtained similarly by removing the {\small $K_{\mathcal{P}(c)i}=0$}.

We define $\mathcal{N}^{(i)}_{\text{pos}}$ and $\mathcal{N}^{(i)}_{\text{neg}}$ as the sets of correctly and incorrectly answered questions by student $i$, respectively. Then we have
\begin{gather}
    \varepsilon_i^{\text{pos}} = \sum_{n: n \in \mathcal{N}^{(i)}_\text{pos}} p_{\bm \theta^{(\tau-1)}}(K_{\mathcal{M}(n)i} = 0|\mathcal{Q}_i)\\
    \varepsilon_i^{\text{neg}} = \sum_{n: n \in \mathcal{N}^{(i)}_\text{neg}} p_{\bm \theta^{(\tau-1)}}(K_{\mathcal{M}(n)i} = 0|\mathcal{Q}_i)\\
    \varepsilon^{(\tau)} = \frac{\sum_i \varepsilon_i^{\text{pos}}}{\sum_i \varepsilon_i^{\text{neg}} + \sum_i \varepsilon_i^{\text{pos}}}.
\end{gather}

For each difficulty class $l$, where $l \in \{\text{easy},\text{medium},\text{hard}\}$, the closed-form solution of the emission probability is
\begin{gather}
    {r_l}_i^{\text{pos}} = \sum_{\substack{n: n \in \mathcal{N}^{(i)}_\text{pos}\\ \text{difficulty}=l}} p_{\bm \theta^{(\tau-1)}}(K_{\mathcal{M}(n)i} = 1|\mathcal{Q}_i)\\
   {r_l}_i^{\text{neg}} = \sum_{\substack{n: n \in \mathcal{N}^{(i)}_\text{neg} \\ \text{difficulty}=l}} p_{\bm \theta^{(\tau-1)}}(K_{\mathcal{M}(n)i} = 1|\mathcal{Q}_i)\\
    {r_l}^{(\tau)} = \frac{\sum_i {r_l}_i^{\text{pos}}}{\sum_i {r_l}_i^{\text{neg}} + \sum_i {r_l}_i^{\text{pos}}}.
\end{gather}

\section{Upward-Downward Algorithm} \label{appx:algorithm}

For each student $i$, we only need to compute the following posterior distributions, which are sufficient for updating all required parameters.
\begin{gather}
p_{\bm \theta}(K_{\mathcal{M}(n)i} = 1|\mathcal{Q}_i) \\
p_{\bm \theta}(K_{\mathcal{M}(n)i} = 0|\mathcal{Q}_i) \\
p_{\bm \theta}(K_{ci} = 1, K_{\mathcal{P}(c)i} = 0|\mathcal{Q}_i) \\
p_{\bm \theta}(K_{ci} = 0, K_{\mathcal{P}(c)i} = 0|\mathcal{Q}_i)
\end{gather}

We first introduce two auxiliary probabilities. The first one is \textbf{downward probability}
\begin{equation}
    \alpha_{c}(k) = p(K_c = k, \mathcal{Q}_{\mathcal{S}_{\mathcal{V} \backslash \mathcal{D}(c)}} = \bm q_{\mathcal{S}_{\mathcal{V} \backslash \mathcal{D}(c)}}).
\end{equation}

Here, $\mathcal{D}(c)$ denotes all the descendant nodes of $c$, \textbf{including} $c$. $\mathcal{V} \backslash \mathcal{D}(c)$ denotes all the nodes except for $\mathcal{D}(c)$. $\mathcal{Q}_{\mathcal{S}_{\mathcal{V} \backslash \mathcal{D}(c)}}$ represents all the questions that involve concepts in $\mathcal{V} \backslash \mathcal{D}(c)$ and $\mathcal{V} \backslash \mathcal{D}(c)$ only. $\mathcal{S}_c$ denotes the set of questions that involves KC $c$ and that is answered by student $i$ (the subscript $i$ is removed for convenience).

The second probability is called the \textbf{upward probability}
\begin{equation}
    \beta_c(k) = p(\mathcal{Q}_{\mathcal{S}_{\mathcal{D}(c)}} = \bm q_{\mathcal{S}_{\mathcal{D}(c)}} | K_c = k).
\end{equation}

The upward probability can be recursively computed from that of the children nodes (denote the children nodes of $c$ as $\textit{Child}(c)$)
\begin{equation}
    \beta_c(k) = \prod_{n \in \mathcal{S}_c}p(Q_n = q_n | K_c = k) \cdot \prod_{j \in \textit{Child}(c)} \tilde{\beta}_{j, c}(k),
\end{equation}
where
\begin{align}
    \tilde{\beta}_{j, \mathcal{P}(j)}(k) &= p(\mathcal{Q}_{\mathcal{S}_{\mathcal{D}(j)}} = \bm q_{\mathcal{S}_{\mathcal{D}(j)}} | K_{\mathcal{P}(j)} = k) \\
    &= \sum_{k_j} \beta_j(k_j) p(K_j = k_j | K_c = k).
\end{align}

At leaf nodes, $\beta_c(k)$ is just a simple emission probability.

The downward probability can be recursively computed from that of the parent node
\begin{align}
    \alpha_{c}(k) &= p(K_c = k, \mathcal{Q}_{\mathcal{S}_{\mathcal{V} \backslash \mathcal{D}(c)}} = \bm q_{\mathcal{S}_{\mathcal{V} \backslash \mathcal{D}(c)}}) \\
       &= \sum_{k_{\mathcal{P}(c)}} p(K_c = k | K_{\mathcal{P}(c)} = k_{\mathcal{P}(c)})\tilde \alpha_{\mathcal{P}(c)},
\end{align}
where
\begin{equation}
\tilde \alpha_{\mathcal{P}(c)}(k) = \alpha_{\mathcal{P}(c)}(k_{\mathcal{P}(c)}) \frac{\beta_{\mathcal{P}(c)}(k_{\mathcal{P}(c)})}{\tilde{\beta}_{c, \mathcal{P}(c)}(k_{\mathcal{P}(c)})}.
\end{equation}

At root nodes, $\alpha_{c}(k)$ is just the prior distribution.

After all the auxiliary variables are computed, the target posterior distributions can be computed as
\begin{equation}
    p_{\bm \theta}(\mathcal{K}_{ci} = 1|\mathcal{Q}_i) = \frac{\alpha_{c}(1)\beta_{c}(1)}{\sum_{k=0}^1 \alpha_{c}(k)\beta_{c}(k)},
\end{equation}

\begin{equation}
\begin{aligned}
    & p_{\bm \theta}(K_{ci} = k_c, K_{\mathcal{P}(c)i} = k_{\mathcal{P}(c)}|\mathcal{Q}_i) \\
    =& \frac{\tilde\alpha_{\mathcal{P}(c)} (k_{\mathcal{P}(c)})\beta_c(k_c)p(k_c | k_{\mathcal{P}(c)})}{\sum_{k'_c, k'_{\mathcal{P}(c)} \in \{0, 1\}} \tilde \alpha_{\mathcal{P}(c)} (k'_{\mathcal{P}(c)})\beta_c(k'_c)p(k'_c |k'_{\mathcal{P}(c)})}.
\end{aligned}
\end{equation}

\section{Translation}\label{appx:translation} 
For both \textsc{XES3G5M} and \textsc{MOOCRadar}, the data translation is done following the pipeline below. 

We translated each question into English using GPT-4o-mini with \textit{[Prompt for Translation]}. We specifically required the LLM to convert all fill-in-the-blank questions into a proper question format (e.g., ‘There are $\underline{\hspace{1em}}$ squares in the plot.’ should be transformed into ‘How many squares are there in the plot?’). Then, the GPT-4o-mini was prompted with \textit{[Prompt for Translation Check]} to self-check the correctness of its translation, considering both the meaning match between the Chinese question and the English translation, as well as the question format conversion. 

We noticed that in the \textsc{XES3G5M} dataset, the blank symbol might be missing in some Chinese fill-in-the-blank questions (e.g., the question ‘There are $\underline{\hspace{1em}}$ squares in the plot.’ might be recorded as ‘There are squares in the plot.’), leading to difficulty in both the translation and the correctness check phases. To handle the incorrect translations, we used a stronger model, GPT-4o, to double-check the translation correctness and generate an explanation for its justification using the same  \textit{[Prompt for Translation Check]}. All questions regarded as incorrect by GPT-4o proceeded into an automatic translation revision phase, where GPT-4o was prompted with the \textit{[Prompt for Translation Fix]} to revise the translation utilizing the explanation of why this translation was incorrect. 

After one round of correction, GPT-4o checked the correctness of the new translation again. After this stage, the remaining incorrect translations were revised by a human translator and added to the collection of English translations. See Appendix~\ref{appx:prompt} for all translation prompts.

\section{Datasets} \label{appx:data-stats}
We summarized the statistics of original \textsc{XES3G5M} and \textsc{MOOCRadar} in Table~\ref{tab:dataset-stats}.

\noindent\textbf{License:} \textsc{XES3G5M} is an open-sourced dataset released under the MIT license, which allow free use of research and educational purposes. \textsc{MOOCRadar} is also publicly available for academic research. Both datasets have been anonymized to protect student privacy and do not contain personal identifiable information.

{\renewcommand{\arraystretch}{1.2}
\begin{table}[h]
  \small
  \centering
  \setlength{\tabcolsep}{4pt}
  \begin{tabular}{lcccc}
    \toprule
    \textbf{Dataset} & \textbf{Students} & \textbf{Items} & \textbf{KCs} & \textbf{Interactions} \\
    \midrule
    XES3G5M   & 18,066 & 7,652  & 865   & 5.5M \\
    MOOCRadar & 14,224 & 2,513  & 5,600 & 12.7M \\
    \hline
  \end{tabular}
  \caption{Statistics of the entire \textsc{XES3G5M} and \textsc{MOOCRadar} datasets. The \textbf{KCs} column here shows the count of fine-grained KCs.}
  \label{tab:dataset-stats}
\end{table}
}

\section{Knowledge Concept Trees} \label{appx:kc-tree-mooc}
In this section, we describe the details of knowledge concept trees for the \textsc{XES3G5M} and \textsc{MOOCRadar}. As noted in Sec.~\ref{sec:data}, \textsc{XES3G5M} provides a predefined hierarchical structure of KCs, which we adopt directly as its knowledge concept tree. In contrast, \textsc{MOOCRadar} includes only fine-grained KCs without any hierarchical organization, so we construct a knowledge concept tree for it.
To begin, we embed the original Chinese KCs using the bge-base-zh model~\cite{bge_embedding}, producing semantic embeddings. These embeddings are then reduced in dimensionality using UMAP~\cite{mcinnes2018umap} and clustered with HDBSCAN~\cite{campello2013density} to group semantically similar KCs. This process yields a set of KC clusters along with a number of outliers that do not belong to any cluster.
For each cluster, we use GPT-4o-mini to generate a summarized KC label. To maintain robustness, we include an ``unsummarizable'' option, allowing the model to flag incoherent clusters, which are then manually reviewed and reclassified.
For the outlier KCs, we sample a representative exercise from the dataset and prompt GPT-4o-mini again to determine whether the KC can be merged into an existing cluster. If not, it is retained as an independent node and manually labeled on a case-by-case basis.

We also apply post-processing to refine the knowledge concept trees. Since we assume each exercise is associated with a single KC, we retain only the most frequently occurring KC when multiple are assigned. Additionally, we merge child nodes associated with fewer than 10 unique exercises into their parent to avoid sparsity. If only one child remains after merging, its questions are reassigned to the parent, and the child node is pruned.

\section{Module Statistics}

Table~\ref{tab:kc-subtree-stats} shows the statistics of all six knowledge modules. The view of each module can be found in Appendix~\ref{appx:modules}. 

{\renewcommand{\arraystretch}{1.2}
\begin{table}[h]
\centering
\small
\setlength{\tabcolsep}{3.5pt}
\begin{tabular}{lcccc}
\toprule
\textbf{Module} & \textbf{\#Nodes} & \textbf{Max Depth} & \textbf{\#Leaves} \\
\midrule
Application Module & 148 & 5 & 100 \\
Computation Module & 95 & 5 & 66  \\
Counting Module & 69 & 5 & 46  \\
Wine Knowledge & 5 & 2 & 4 \\
Circuit Design & 9 & 2 & 8 \\
Education Theory & 4 & 2 & 3 \\
\bottomrule
\end{tabular}
\caption{Statistics of the six knowledge modules sampled from \textsc{XES3G5M} and \textsc{MOOCRadar}.}
\label{tab:kc-subtree-stats}
\end{table}
}

\section{Implementation Details}

We initialize our model parameters as in Table~\ref{tab:param-init}. 

{\renewcommand{\arraystretch}{1.2}
\begin{table}[h]
\small
\centering
\begin{tabular}{lcc}
\toprule
\textbf{Parameter} & \textbf{Symbol} & \textbf{Initial Value} \\
\midrule
Transition & $\gamma_c$ & 0.1\\
Emission (easy) & $r_\text{easy}$ & 0.9\\
Emission (medium) & $r_\text{med}$ & 0.8\\
Emission (hard) & $r_\text{hard}$ & 0.75\\
Guessing & $\varepsilon$ & 0.1\\
\hline
\end{tabular}
\caption{Initialization values for model parameters before the Expectation-Maximization estimation.}
\label{tab:param-init}
\end{table}
}

We observe that the guessing probability $\varepsilon$ can exceed $0.3$ during training in some cases. To avoid overfitting to spurious correct responses, we clip $\varepsilon$ to a maximum of $0.3$ in each EM update iteration. This constraint prevents $\varepsilon$ from approaching the emission probability of hard questions, which would make it difficult to distinguish between true mastery and random guessing.

For all deep-learning-based baseline methods, we follow the default hyperparameters reported in their original papers to train the offline versions. For their online variants, we set the learning rate to $1e-4$ and the batch size to $32$, and train for a single epoch at each update step.

For both LLM-based baseline methods, we use the vLLM package~\cite{kwon2023efficient} for inference, with a sampling temperature of 0.8, a top-p (nucleus) sampling threshold of 0.8, and a repetition penalty of 1.1.

\clearpage
\onecolumn
\section{Module Views} \label{appx:modules}
\begin{center}
\footnotesize 
\begin{Verbatim}[fontsize=\footnotesize]
Application Module
    +-- Equation Word Problem
    |   +-- Linear Equation Word Problem
    |   \-- Indeterminate Equation Word Problem
    +-- Addition/Subtraction Word Problem
    |   +-- Change in Ratio Word Problem
    |   \-- Simultaneous Increase/Decrease Word Problem
    +-- Period Problem
    |   +-- Basic Arrangement Period Problem
    |   +-- Sequence Operation Period Problem
    |   |   +-- Number Period
    |   |   \-- Find Which Number in Sequence Problem
    |   +-- Period Problem in Time
    |   |   +-- Period in Calendar Dates
    |   |   +-- Period in Time
    |   |   \-- Find Day of the Week for a Date Problem
    |   \-- Cyclic Operation Period Problem
    +-- Sum and Difference Multiple Problem
    |   +-- Basic Calculation of Multiples
    |   |   +-- Two Quantities Multiple
    |   |   \-- Multiple Quantities Multiple
    |   +-- Change Multiple Problem
    |   |   +-- Two Quantities Simultaneously Change Multiple
    |   |   \-- Basic Change Multiple
    |   +-- Sum and Multiple Problem
    |   |   +-- Two Quantities Sum and Multiple Problem
    |   |   |   \-- Two Quantities Sum and Multiple
    |   |   +-- Multiple Quantities Sum and Multiple Problem
    |   |   |   \-- Multiple Quantities Sum and Multiple
    |   |   +-- Find Hidden Sum
    |   |   \-- From Less to More
    |   +-- Sum and Difference Problem
    |   |   +-- Two Quantities Sum and Difference Problem
    |   |   |   +-- Known Differences
    |   |   |   +-- Known and Hidden Difference
    |   |   |   \-- Hidden and Known Difference
    |   |   \-- Multiple Quantities Sum and Difference Problem
    |   \-- Difference Multiple
    |       +-- Two Quantities Difference Multiple
    |       |   +-- Integer Ratio with Difference
    |       |   +-- Hidden Ratio Type Two Quantities Difference Multiple Problem
    |       |   +-- Hidden Difference Type Two Quantities Difference Multiple Problem
    |       |   +-- Non-integer Ratio Difference Multiple Insufficient
    |       |   \-- Non-integer Ratio Difference Multiple Remainder
    |       \-- Multiple Quantities Difference Multiple
    
    ...
    
    \-- Chicken-Rabbit Problem
        +-- Assumption Method for Solving Chicken-Rabbit Problem
        |   +-- Back-Deduction Type
        |   +-- Basic Type
        |   |   +-- Prototype Problem
        |   |   \-- Variant Problems
        |   \-- Find Number of Animals
        +-- Grouping Method for Solving Chicken-Rabbit Problem
        |   +-- Head Multiple Type
        |   \-- Flexible Grouping
        +-- Multiple Quantities Chicken-Rabbit Problem
        +-- Using Assumption Method to Solve Modified Chicken-Rabbit Problem
        +-- Using Grouping Method to Solve Modified Chicken-Rabbit Problem
        \-- Using Grouping Method to Solve Basic Chicken-Rabbit Problem
\end{Verbatim}
\end{center}

\clearpage
\begin{center}
\footnotesize 
\begin{Verbatim}[fontsize=\footnotesize]
Computation Module
    +-- Exponentiation
    |   +-- Application of Exponentiation
    |   \-- Exponentiation Operations
    +-- Fraction
    |   +-- Fraction Basics
    |   |   +-- Properties of Fractions
    |   |   \-- Meaning of Fractions
    |   +-- Fraction Tricks
    |   \-- Fraction Operations
    |       +-- Fraction Addition and Subtraction
    |       \-- Addition and Subtraction of Fractions with Different Denominators
    +-- Unit Conversion
    |   +-- Length Unit Conversion
    |   \-- Area Unit Conversion
    +-- Define New Operation
    |   +-- Reverse Solving Unknown Type
    |   \-- Direct Calculation Type
    |       \-- Normal Type
    +-- Decimal
    |   +-- Decimal Addition and Subtraction
    |   |   +-- Decimal Addition and Subtraction Trick with Rounding Method
    |   |   \-- Decimal Addition and Subtraction Vertical Format Calculation
    |   +-- Decimal Four Operations
    |   +-- Decimal Basics
    |   |   +-- Rounding
    |   |   +-- Decimal Comparison
    |   |   +-- Decimal Point Movement Patterns
    |   |   \-- Understanding Decimals
    |   \-- Meaning of Decimals
    |       +-- Decimal Point Movement
    |       \-- Reading and Writing Decimals
    +-- Sequences and Number Tables
    |   +-- Sequence Patterns
    |   +-- Number Table Patterns
    |   |   +-- Finding Patterns by Combining Numbers and Diagrams (Multiple Diagrams)
    |   |   \-- Rectangle Number Table
    |   |       +-- Positional Relationship
    |   |       +-- Find Number at Known Position in Continuous Natural Number Rectangle Table
    |   |       \-- Find Position of Known Number in Continuous Natural Number Rectangle Table
    |   +-- Arithmetic Sequence
    |   |   +-- Application of Mean Value Theorem
    |   |   +-- Truncated Sum of Arithmetic Sequence
    |   |   +-- Find Common Difference of Arithmetic Sequence
    |   |   +-- Sum of Arithmetic Sequence
    |   |   +-- Find General Term of Arithmetic Sequence
    |   |   \-- Find Number of Terms in Arithmetic Sequence
    |   \-- Geometric Sequence
     
    ...
    
    +-- Equation Basics
    |   +-- Linear Equation in One Variable
    |   |   \-- Equation with Integer Coefficients
    |   \-- Indeterminate Equation
    +-- Comparison and Estimation
    \-- Induction of Split and General Terms
\end{Verbatim}
\end{center}

\clearpage
\begin{center}
\footnotesize 
\begin{Verbatim}[fontsize=\footnotesize]
Counting Module
    +-- Geometric Counting
    |   +-- Categorical Enumeration of Figures
    |   |   +-- Regular Figure Enumeration Counting
    |   |   |   +-- Categorized Figures
    |   |   |   +-- Square
    |   |   |   +-- Understanding Line Segments
    |   |   |   \-- Rectangle
    |   |   \-- Lattice Point Constructed Figures
    |   +-- Correspondence Method Figures
    |   |   \-- Counting by Multiplication
    |   |       +-- Count Triangles
    |   |       +-- Count Lines
    |   |       \-- Count Rectangles
    |   \-- Auxiliary Line
    +-- Addition-Multiplication Principle
    |   +-- Multiplication Principle
    |   |   +-- Other Types in Multiplication Principle
    |   |   +-- Object Quantity Combination
    |   |   +-- Item Matching
    |   |   +-- Item Matching (With Special Requirements)
    |   |   \-- Route Matching Problem
    |   +-- Comprehensive Addition-Multiplication Principle
    |   +-- Addition Principle
    |   |   +-- Other Types in Addition Principle
    |   |   \-- Handshake and Toasting Problem
    |   +-- Queueing Problem
    |   +-- Coloring Counting Problem
    |   \-- Grouping Problem
    |       +-- General Grouping Problem
    |       \-- Grouping Problem with Special Requirements
    +-- Inclusion-Exclusion Principle
    |   +-- Three-Set Inclusion-Exclusion
    |   +-- Two-Set Inclusion-Exclusion
    |   \-- Geometric Counting with Multi-Set Inclusion-Exclusion
    +-- Permutation and Combination
    |   +-- Permutation
    |   |   +-- Optimal Limitation Method
    |   |   \-- Basic Application of Permutation
    |   +-- Permutation Number
    |   \-- Combination

    ...
    
    +-- Counting Method
    |   +-- Standard Counting Method
    |   |   \-- Shortest Path
    |   +-- Special Points or Areas
    |   \-- Stepwise Counting Method
    \-- Statistics and Probability
        +-- Probability
        \-- Statistical Charts
\end{Verbatim}
\end{center}

\clearpage
\begin{center}
\footnotesize 
\begin{Verbatim}[fontsize=\footnotesize]
Wine Knowledge
    +-- Wine
    +-- Wine Evaluation Knowledge
    +-- Wine Aroma Type
    \-- Types of Wine

Circuit Design
    +-- Amplifier Circuit Design
    +-- Ideal Operational Amplifier
    +-- Electronics
    +-- Electrical Concepts
    +-- Electrical Circuit Knowledge
    +-- Circuit Design and Analysis
    +-- Operational Amplifier Knowledge
    \-- Integrated Operational Amplifier Circuit

Education Theory
    +-- Feedback
    +-- Class Teacher Management and Teacher-Student Relationship
    \-- Foundation of Class Formation
\end{Verbatim}
\end{center}

\newpage
\section{LLM Prompt} \label{appx:prompt}
In this section, we provide the complete prompts for LLM baselines and dataset translation.
\begin{tcolorbox}[left=1.2pt,right=1.2pt,top=1.2pt,bottom=1.2pt]
\small
\textbf{Prompt for Translation:}
\\
\\
You are a helpful AI assistant skilled in translating Chinese exercises to English.
\\
\\
Guidelines:\\
- You will be provided with an exercise in Chinese. Your task is to translate it accurately and clearly into English.\\
- If the Chinese question is a fill-in-the-blank question, convert it into a proper question format in English. Be mindful that the blank symbol might be missing from the original question due to formatting errors.\\
- Output only the translated English text. Do not include any additional text, explanations, or formatting beyond the translation.
\\
\\
Examples:
\\
User 1: \\
Here is the math exercise to translate: \textit{(Omit the Chinese question)}\\
Assistant 1:\\
Xiao Ming has $10$ apples, Xiao Hong has $5$ apples, how many more apples does Xiao Ming have than Xiao Hong?
\\
...
\\
\textit{(Omit other 2 examples)}
\\ 
\\
Here is the math exercise to translate: \textit{(Omit the Chinese question)}
\end{tcolorbox}

\begin{tcolorbox}[left=1.2pt,right=1.2pt,top=1.2pt,bottom=1.2pt]
\small
\textbf{Prompt for Translation Check:}
\\
\\
You are a helpful AI assistant skilled in assessing the quality of Chinese question translations for accuracy and coherence. You will be provided with two versions of an exercise: one in Chinese and one in English. Your task is to evaluate whether the English translation accurately reflects the meaning and intent of the Chinese question and provide a detailed explanation to justify your assessment.
\\
Guidelines:\\
- If the Chinese question is a fill-in-the-blank question, the English translation must be rewritten into a proper question format without retaining the blank symbol. If it fails to meet this requirement, consider it an incorrect translation.\\
- Be mindful of potential formatting errors in the Chinese question. For example, the blank symbol in a fill-in-the-blank question might be missing. Carefully discern whether the question is a statement or a fill-in-the-blank question with a missing blank.\\
- If the Chinese question references an image, it is acceptable for the English translation to omit the image filename as long as it maintains the question's meaning.\\
- Output your evaluation in JSON format using the provided template. First, explain why the English translation is correct or incorrect, then provide a final justification as a boolean value (True for correct, False for incorrect).\\
- Do not include any additional text or explanations beyond the required JSON output.\\
\\
\\
Template:\\
\{\\
"explanation": (Your explanation on why you think the English translation is correct/wrong),\\
"correct\_translation": (a boolean value, True if the English translation is a correct translation, False otherwise)\\
\}
\\
\\
Examples:
\\
User 1:\\
Here is the Chinese question:\textit{(Omit the Chinese question)}\\
Here is the English translation of the question: A number, when divided by $4$, added $4$, multiplied by $4$, and then subtracted $4$, results in $16$. What is the number?\\
Assistant 1: \\
\{\\
"explanation": "The Chinese question describes a sequence of mathematical operations performed on a number, leading to the result of 16. The English translation accurately conveys the meaning of the Chinese question by describing the same sequence of operations and the expected result, while appropriately converting the blank into the question format `What is the number?'. Both the meaning and format are correct.", \\
"correct\_translation": \\
\}
\\
...
\\
\textit{(Omit other 2 examples)}
\\ 
\\
Here is the Chinese question:\textit{(Omit the Chinese question)}\\
Here is the English translation of the question:\textit{(Omit the English translation)}
\end{tcolorbox}

\begin{tcolorbox}[left=1.2pt,right=1.2pt,top=1.2pt,bottom=1.2pt]
\small
\textbf{Prompt for Translation Fix:}
\\
\\
You are a helpful AI assistant skilled in improving the translation of Chinese questions to ensure accuracy and coherence. You will be provided with two versions of a question: one in Chinese and one in English, along with an explanation of why the translation is incorrect. Your task is to rewrite the English translation based on the given explanation to make it correct and consistent with the original Chinese question.
\\
\\
Guidelines:\\
- Be mindful of potential formatting errors in the Chinese question. For example, the blank symbol in a fill-in-the-blank question may be missing. Carefully discern whether the question is a statement or a fill-in-the-blank question with a missing blank. \\
- If the Chinese question is a fill-in-the-blank question, rewrite the English translation as a proper question without retaining the blank symbol.\\
- Provide only the corrected English translation as the output. Avoid including additional explanations or text.
\\
\\
Examples:
\\
User 1: \\
Here is the Chinese question:\textit{(Omit the Chinese question)}\\
Here is the English translation you should rewrite: A number, when divided by $4$, added $4$, multiplied by $4$, and then subtracted $4$, results in $16$. Then the number is ()\\
The reason why the translation is incorrect:
The Chinese question is a fill-in-the-blank question as indicated by the blank symbol (), which requires the English translation to be reformatted into a proper question format without the blank. The provided English translation retains the blank, which does not conform to the specified criteria for a correct translation.\\
Assistant 1:\\
Xiao Ming has $10$ apples, Xiao Hong has $5$ apples, how many more apples does Xiao Ming have than Xiao Hong?
\\
\\
...
\\
\textit{(Omit other 2 examples)}
\\ 
\\
Here is the math exercise to translate: \textit{(Omit the Chinese question)}\\
Here is the English translation you should rewrite:\textit{(Omit the English translation)}\\
The reason why the translation is incorrect: \textit{(Omit the explanation)}
\end{tcolorbox}

\begin{tcolorbox}[left=1.2pt,right=1.2pt,top=1.2pt,bottom=1.2pt]
\small
\textbf{Prompt for LLM Knowledge Tracing:}
\\
\\
Your task is to analyze student's past performance on a series of questions and predict whether students can answer a given question correctly. Below are 10 questions the student has already answered. After each question you will see whether the answer was correct (1) or incorrect (0).
\\
\\
Q1. In a capacitive coupling amplifier circuit, after introducing negative feedback, it is only possible to have low-frequency self-excited oscillation, and high-frequency self-excited oscillation is not possible. Is this statement true or false?\\
Concepts: Feedback\\
Correct: 0\\
\\
Q2. Since some negative feedback amplifiers can produce self-excited oscillations, can they be used as signal sources?\\
Concepts: Feedback\\
Correct: 0\\
\\
Q3. ... \\
\textit{(Omit the other 8 historical exercises)}\\
\\
Now consider the new question:\\
If negative feedback is introduced through a resistor in a single-stage common-emitter amplifier, what will happen? If negative feedback is introduced through a resistor in a two-stage common-emitter amplifier, what will happen? A. It will definitely produce high-frequency self-oscillation B. It may produce high-frequency self-oscillation C. It will definitely not produce high-frequency self-oscillation.\\
Concepts: Feedback\\
Based on the student's past performance, will the student answer this question correctly? Respond with 1 for correct, 0 for incorrect. Output only 0, 1—no additional text.\\
\\Predict:
\end{tcolorbox}

\end{document}